\documentclass[12pt]{article} 
\usepackage{times}   
\usepackage{helvet}  
\usepackage{courier} 
\usepackage{url}     
\usepackage{natbib}     
\usepackage{fullpage}     
\usepackage{graphicx}
\usepackage{subcaption}
\usepackage{float}

\usepackage{booktabs}       
\usepackage{microtype}      
\usepackage{nicefrac}       
\usepackage{array}
\newcolumntype{P}[1]{>{\centering\arraybackslash}p{#1}}
\usepackage[usenames, dvipsnames]{color}

\usepackage{amsfonts}        
\usepackage{amsmath,amssymb} 
\usepackage{mathtools}

\usepackage{algorithm}
\usepackage[noend]{algpseudocode}

\usepackage{graphicx}
\graphicspath{ {images/} }
\usepackage{authblk}

\title{How do Humans Understand Explanations from Machine Learning Systems? \\ An Evaluation of the Human-Interpretability of Explanation}

\author[1]{Menaka Narayanan*}
\author[1]{Emily Chen*}
\author[1]{Jeffrey He*}
\author[2]{Been Kim} 
\author[1]{Sam Gershman} 
\author[1]{Finale Doshi-Velez} 
\affil[1]{Harvard University}
\affil[2]{Google Brain}

\begin{document}

\maketitle

\begin{abstract}
Recent years have seen a boom in interest in machine learning systems
that can provide a human-understandable rationale for their predictions
or decisions.  However, exactly what kinds of explanation are truly
human-interpretable remains poorly understood.  This work advances our
understanding of what makes explanations interpretable in the specific
context of verification. Suppose we have a machine learning system that predicts X, and we provide rationale for this prediction X. Given an input, an explanation, and an
output, is the output consistent with the input and the supposed
rationale?  Via a series of user-studies, we identify what kinds of
increases in complexity have the greatest effect on the time it takes for humans to verify the rationale, and
which seem relatively insensitive.  
\end{abstract}

\section{Introduction}
\label{sec:intro}

Interpretable machine learning systems provide not only decisions or
predictions but also explanation for their outputs. Explanations can
help increase trust and safety by identifying when the recommendation
is reasonable when it is not.  While interpretability has a long
history in AI \citep{michie1988machine}, the relatively recent
widespread adoption of machine learning systems in real, complex
environments has lead to an increased attention to interpretable
machine learning systems, with applications including understanding
notifications on mobile devices
\citep{mehrotra2017interpretable,wang2016bayesian}, calculating stroke
risk \citep{letham2015interpretable}, and designing materials
\citep{raccuglia2016machine}.  Techniques for ascertaining the
provenance of a prediction are also popular within the machine
learning community as ways for us to simply understand our
increasingly complex models
\citep{lei2016rationalizing,selvaraju2016grad,adler2016auditing}.

The increased interest in interpretability has resulted in many forms
of explanation being proposed, ranging from classical approaches
such as decision trees \citep{breiman1984classification} to input
gradients or other forms of (possibly smoothed) sensitivity analysis
\citep{selvaraju2016grad,ribeiro2016should,lei2016rationalizing},
generalized additive models \citep{caruana2015intelligible},
procedures \citep{singh2016programs}, falling rule lists
\citep{wang2015falling}, exemplars
\citep{kim2014bayesian,frey2007clustering} and decision sets
\citep{lakkaraju2016interpretable}---to name a few.  In all of these
cases, there is a face-validity to the proposed form of explanation:
if the explanation was not human-interpretable, clearly it would not
have passed peer review.

That said, these works provide little guidance about when different
kinds of explanation might be appropriate, and within a class of
explanations---such as decision-trees or decision-sets---what are the
limitations of human reasoning.  For example, it is hard to imagine
that a human would find a 5000-node decision tree as interpretable as
5-node decision tree, for any reasonable notion of interpretable.  The
reason the explanation is desired is also often left implicit.  In
\citet{doshi2017roadmap}, we point to a growing need for the
interpretable machine learning community to engage with the human
factors and cognitive science of interpretability: we can spend
enormous efforts optimizing all kinds of models and regularizers, but
that effort is only worthwhile if those models and regularizers
actually solve the original human-centered task of providing
explanation.  Carefully controlled human-subject experiments provide
an evidence-based approach to identify what kinds of regularizers we
should be using.

In this work, we make modest but concrete strides toward this large
goal of quantifying what makes explanation human-interpretable.  We
shall assume that there exists some explanation system that generates
the explanation---for example, there exist a variety of approaches
that use perturbations around a point of interest to produce a local
decision boundary \citep{ribeiro2016should,singh2016programs}.  Our
question is: What kinds of explanation can humans most easily utilize?
Below, we describe the kind of task we shall explore, as well as the
form of the explanation.

\paragraph{Choice of Task}
The question of what kinds of explanation a human can utilize implies
the presence of a downstream task.  The task may be
intrinsic---relying on just the explanation alone---or
extrinsic---relying on the explanation and other facts about the
environment.\footnote{As noted in \citet{herman2017promise},
  explanation can also be used to persuade rather than inform; we
  exclude that use-case here.}  Intrinsic tasks include problems such
as verification---given an input, output, and explanation, can the
human user verify that the output is consistent with the input and
provided explanation?---and counterfactual reasoning---given an input,
output, and explanation, can the human subject identify what input
changes would result in a different output?  In contrast, extrinsic
tasks include goals such as safety---given the input, output,
explanation, and observations of the world, does the explanation help
the human user identify when the agent is going to make a
mistake?---and trust---given the input, output, explanation, and
observations of the world, does the explanation increase the human
user's confidence in the agent?  Evaluation on extrinsic tasks, while
ultimately what we care about, require careful experimental design
that ensures all subjects have similar knowledge and assumptions about
the environment.  One must also tease apart conflations between human
perception of knowledge with actual knowledge: for example, it may be
possible to manipulate trust arbitrarily separately from prediction
accuracy.  Thus, evaluating extrinsic tasks is more challenging than
intrinsic ones.

In this work, we will focus on the simplest intrinsic setting:
verification.  Given a specific input, explanation, and output, can we
quickly determine whether the output is consistent with the input and
explanation?  Such a setting may arise in consumer recommendation
scenarios---for example, a salesperson may wish to ensure that a
specific product recommendation is consistent with a consumer's
preferences.  Starting simple also provides an opportunity to explore
aspects relevant to the experimental design.

\paragraph{Choice of Explanation Form}
As mentioned above, there have been many forms of explanation
proposed, ranging from decision trees to gradients of neural networks.
In this work, we consider explanations that are in the form of
\emph{decision sets}.  Decision sets are a particular form of
procedure consisting of a collection of cases, each mapping some
function of the inputs to a collection of outputs.  An example of a 
decision set is given below
\begin{figure}[H] 
  \centering
  \includegraphics[width=3in]{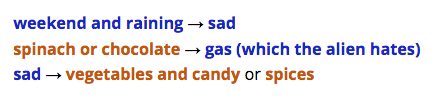}
  \caption{Example of a decision set explanation.}
  \label{fig:rule_set}
\end{figure}
\noindent
where each line contains a clause in disjunctive normal form (an
or-of-ands) of the inputs, which, if true, provides a way to verify
the output (also in disjunctive normal form).  As argued in
\citet{lakkaraju2016interpretable}, decision sets are relatively easy
for humans to parse given a specific instance, because they can scan
for the rule that applies and choose the accompanying output.
Decision sets (also known as rule sets) also enjoy a long history of
optimization techniques, ranging from \citet{frank1998generating,
  cohen1995fast, clark1991rule} to \citet{lakkaraju2016interpretable}.

However, we can also already see that there are many factors that
could potentially make the decision set more or less challenging to
follow: in addition to the number of lines, there are ways in which
terms interact in the disjunctions and conjunctions, and, more subtly,
aspects such as how often terms appear and whether terms represent
intermediate concepts.  Which of these factors are most relevant when
it comes to a human's ability to process and utilize an explanation,
and to what extent?  The answer to this question has important
implications for the design of explanation systems in interpretable
machine learning, especially if we find that our
explanation-processing ability is relatively robust to variation in
some factors but not others.

\paragraph{Contributions}
The core contribution of this work is to provide an empirical
grounding for what kinds of explanations humans can utilize.  We find
that while almost all increases in the complexity of an explanation
result in longer response times, some types of complexity---such as
the number of lines, or the number of new concepts introduced---have a
much bigger effect than others---such as variable repetition.  We also
find some unintuitive patterns, such as how participants seem to prefer
an explanation that requires a single, more complex line to one that
spans multiple simpler lines (each defining a new concept for the next
line).  While more work is clearly needed in this area, we take
initial steps in identifying what kinds of factors are most important
to optimize for when providing explanation to humans.

\section{Related Work}

Interpretable machine learning methods aim to optimize models for both
succinct explanation and predictive performance.  Common types of
explanation include regressions with simple, human-simulatable
functions \citep{caruana2015intelligible, kim2015ibcm,
  ruping2006learning,buciluǎ2006model, ustun2016supersparse,
  doshi2014graph, kim2015mind, krakovna2016increasing,
  hughes2016supervised, jung2017simple}, various kinds of logic-based
methods
\citep{wang2015falling,lakkaraju2016interpretable,singh2016programs,liu2016sparse,safavian1991survey,
  bayesian2017wang}, techniques for extracting local explanations from
black-box models
\citep{ribeiro2016should,lei2016rationalizing,adler2016auditing,selvaraju2016grad,
  smilkov2017smoothgrad,shrikumar2016not,kindermans2017patternnet,
  ross2017right}, and visualization \citep{wattenberg2016attacking}.
There exist a range of technical approaches to derive each form of
explanation, whether it be learning sparse models
\citep{mehmood2012review,chandrashekar2014survey}, monotone functions
\citep{canini2016fast}, or efficient logic-based models
\citep{rivest1987learning}.  Related to our work, there also exists a
history of identify human-relevant concepts from data, including
disentangled representations \citep{chen2016infogan} and predicate
invention in inductive logic programming \citep{muggleton2015meta}.
While the algorithms are sophisticated, the measures of
interpretability are often not---it is common for researchers to
simply appeal to the face-validity of the results that they find
(i.e., ``this result makes sense to the human reader'')
\citep{caruana2015intelligible,lei2016rationalizing,ribeiro2016should}.

In parallel, the literature on explanation in psychology also offers
several general insights into the design of interpretable AI
systems. For example, humans prefer explanations that are both simple
and highly probable \citep{lombrozo07}. Human explanations typically
appeal to causal structure \citep{lombrozo2006structure} and
counterfactuals \citep{keil2006explanation}.
\citet{miller1956magical} famously argued that humans can hold about
seven items simultaneously in working memory, suggesting that
human-interpretable explanations should obey some kind of capacity
limit (importantly, these items can correspond to complex
\emph{cognitive chunks}---for example, `CIAFBINSA' is easier to
remember when it is chunked as `CIA', `FBI', `NSA.').  Orthogonally,
\citet{kahneman2011thinking} notes that humans have different modes of
thinking, and larger explanations might push humans into a more
careful, rational thinking mode.  Machine learning researchers can
convert these concepts into notions such as sparsity or
simulatability, but the work to determine answers to questions such as
``how sparse?'' or ``how long?'' requires empirical evaluation.  

Existing studies evaluting the human-interpretability of explanation
often fall into the A-B test framework, in which a proposed model is
being compared to some competitor, generally on an intrinsic task.
For example, \citet{kim2014bayesian} showed that human subjects'
performance on a classification task was better when using examples as
representation than when using non-example-based
representation. \citet{lakkaraju2016interpretable} performed a user
study in which they found subjects are faster and more accurate at
describing local decision boundaries based on decision sets rather
than rule lists.  \citet{subramanian1992comparison} found that users
prefer decision trees to tables in games, whereas
\citet{huysmans2011empirical} found users prefer, and are more
accurate, with decision tables rather than other classifiers in a
credit scoring domain.  \citet{hayete2004gotrees} found a preference
for non-oblique splits in decision trees (see
\citet{freitas2014comprehensible} for more detailed survey).  These
works provide quantitative evaluations of the human-interpretability
of explanation, but rarely identify what properties are most essential
for what contexts---which is critical for generalization.

Specific application areas have also evaluated the desired properties
of an explanation within the context of the application.  For example,
\citet{tintarev2015explaining} provides a survey in the context of
recommendation systems, noting differences between the kind of
explanations that manipulate trust \citep{cosley2003seeing} and the
kind that increase the odds of a good decision
\citep{bilgic2005explaining}.  In many cases, these studies are
looking at whether the explanation has an effect, sometimes also
considering a few different kinds of explanation (actions of similar
customers, etc.).  \citet{horsky2012interface} describe how presenting
the right clinical data alongside a decision support recommendation
can help with adoption and trust.  \citet{bussone2015role} found that
overly detailed explanations from clinical decision support systems
enhance trust but also create over-reliance; short or absent
explanations prevent over-reliance but decrease trust.  These studies
span a variety of extrinsic tasks, and again given the specificity of
each explanation type, identifying generalizable properties is
challenging.

Closer to the objectives of the proposed work, \citet{kulesza2013too}
performed a qualitative study in which they varied the soundness
(nothing but the truth) and the completeness (the whole truth) of an
explanation in a recommendation system setting.  They found
completeness was important for participants to build accurate mental
models of the system.  \citet{allahyari2011user,elomaa2017defense}
also find that larger models can sometimes be more interpretable.
\citet{schmid2016does} find that human-recognizable intermediate
predicates in inductive knowledge programs can sometimes improve
simulation time.  \citet{sangdeh2017manipulating} manipulate the size
and transparency of an explanation and find that longer explanations
and black-box models are harder to simulate accurately (even given
many instances) on a real-world application predicting housing prices.
Our work fits into this category of empirical study of explanation
evaluation; we perform controlled studies on a pair of synthetic
application to assess the effect of a large set of explanation
parameters.



\section{Methods}
Our main research question is to determine what properties of decision
sets are most relevant for human users to be able to utilize the
explanations for verification.  In order to carefully control various
properties of the explanation and the context, in the following we
shall present human subjects with explanations that \emph{could} have
been machine-generated, but were in fact generated by us.  Before
describing our experiment, we emphasize that while our explanations
are not actually machine-generated, our findings provide suggestions
to designers of interpretable machine learning systems about what
parameters affect the usability of an explanation, and which should be
optimized when producing explanations.

\subsection{Factors Varied}
Even within decision sets, there are a large number of ways in which
the explanations could be varied.  Following initial pilot studies
(see Appendix), we chose to focus on the three main kinds of variation
(described below).  We also tested on two different domains---a faux
recipe recommendation domain and a faux clinical decision support
domain---to see if the context would result in different explanation
processing while other factors were held constant.

\paragraph{Explanation Variation}
We explored the following sources of explanation variation: 

\begin{itemize}

\item \textbf{V1: Explanation Size.}  We varied the size of the explanation across two dimensions: the \emph{total number of lines} in the decision set, and the \emph{maximum number of terms within the output clause}.  The first corresponds to increasing the vertical size of the explanation---the number of cases---while the second corresponds to increasing the horizontal size of the explanation---the complexity of each case.  We focused on output clauses because they were harder to parse: input clauses could be quickly scanned for setting-related terms, but output clauses had to be read through and processed completely to verify an explanation.  We hypothesized that increasing the size of the explanation across either dimension would increase the time required to perform the verification task.

\item \textbf{V2: Creating New Types of Cognitive Chunks.}
In Figure~\ref{fig:interface}, the first line of the decision set introduces a new cognitive chunk: if the alien is `checking the news' or `coughing,' that corresponds to a new concept `windy.'\footnote{One can imagine this
    being akin to giving names to nodes in a complex model, such as a
    neural network.}  On one hand, creating new cognitive chunks can make an explanation more succinct.  On the other hand, the human must now process an additional idea.  We varied two aspects related to new cognitive chunks.  First, we simply adjusted the number of new cognitive chunks present in the explanation.  All of the cognitive chunks were necessary for verification, to ensure that the participant had to traverse all of concepts instead of skimming for the relevant one.

Second, we tested whether it was more helpful to introduce a new cognitive chunk or leave it implicit: for example, instead of introducing a concept `windy' for `checking the news or coughing,' (explicit) we could simply include `checking the news or coughing' wherever windy appeared (implicit).  We hypothesized that adding cognitive chunks would increase the time required to process an explanation, because the user would have to consider more lines in the decision list to come to a conclusion.  However, we hypothesized that it would still be more time-efficient to introduce the new chunk rather than having long clauses that implicitly contained the meaning of the chunk.  
   
\item \textbf{V3: Repeated Terms in an Explanation.} Another factor that might affect humans' ability to process an explanation is how often terms are used.  For example, if input conditions in the decision list have little overlap, then it may be faster to find the appropriate one because there are fewer relevant cases to consider.  We hypothesized that if an input condition appeared in several lines of the explanation, this would increase the time it took to search for the correct rule in the explanation.  (Repeated terms was also a factor used by \cite{lakkaraju2016interpretable} to measure interpretability.)

\end{itemize}

\paragraph{Domain Variation} 
Below we describe the two contexts, or domains, that we used in our
experiments: recipe recommendations (familiar) and clinical decision
support (unfamiliar).  The domains were designed to feel very
different but such that the verification tasks could be made to
exactly parallel each other, allowing us to investigate the effect of
the domain context in situations when the form of the explanation was
exactly the same.  We hypothesized that these trends would be
consistent in both the recipe and clinical domains.

\emph{Alien Recipes.}  In the first domain, study participants were
told that the machine learning system had studied a group of aliens
and determined each of their individual food preferences in various
settings (e.g., snowing, weekend).  Each task involved presenting
participants with the setting (the input), the systems's description
of the current alien's preferences (the explanation), and a set of
recommended ingredients (the output).  The user was then asked whether
the ingredients recommendation was a good one.  This scenario
represents a setting in which customers may wish to know why a certain
product or products were recommended to them.  Aliens were introduced
in order to avoid the subject's prior knowledge or preferences about
settings and food affecting their responses; each task involved a
different alien so that each explanation could be unique.  All
non-literals (e.g. what ingredients were spices) were defined in a
dictionary so that all participants would have the same cognitive
chunks.

\emph{Alien Medicine} In the second domain, study participants were
told that the machine learning system had studied a group of aliens
and determined personalized treatment strategies for various symptoms
(e.g. sore throat).  Each task involved presenting participants with
the symptoms (the input), the system's description of the alien's
personalized treatment strategy (the explanation), and a set of
recommended drugs (the output).  The user was then asked whether the
drugs recommended were appropriate.  This scenario closely matches a
clinical decision support setting in which a system might suggest a
treatment given a patient's symptoms, and the clinician may wish to
know why the system chose a particular treatment.

As before, aliens were chosen to avoid the subject's prior medical
knowledge from affecting their responses; each task involved
personalized medicine so that each explanation could be unique.  We
chose drug names that corresponded with the first letter of the
illness (e.g. antibiotic medications were Aerove, Adenon and Athoxin)
so as to replicate the level of ease and familiarity of food names.
Again, all drug names and categories were provided in a dictionary so
that participants would have the same cognitive chunks.

In our experiments, we maintained an exact correspondence between
inputs (setting vs. symptoms), outputs (foods vs. drugs), categories
(food categories vs. drug categories), and the forms of the
explanation.  These parallels allowed us to test whether changing the
domain from a relatively familiar, low-risk product recommendation
setting to a relatively unfamiliar, higher-risk decision-making
setting affected how subjects utilized the explanations for
verification.

\subsection{Experimental Design and Interface}
The three kinds of variation and two domains resulted in six total
experiments.  In the recipe domain, we held the list of ingredients,
food categories, and possible input conditions constant.  Similarly,
in the clinical domain, we held the list of symptoms, medicine
categories, and possible input conditions constant.  The levels were
as follows:
\begin{itemize}
  
\item Length of the explanation (V1).  We manipulated the length of
  the explanation (varying between 2, 6, and 10 lines) and the length
  of the output clause (varying between 2 and 5 terms).  Each
  combination was tested twice within a experiment, for a total of 12
  questions.

\item Introducing new concepts (V2): We manipulated the number of
  cognitive chunks introduced (varying from 1 to 5), and whether they
  were embedded into the explanation or abstracted out into new
  cognitive chunks.  Each combination was tested once within a
  experiment, for a total of 10 questions.  

\item Repeated terms (V3): We manipulated the number of times the
  input conditions appeared in the explanation (varying from 1 to 5)
  and held the number of lines and length of clauses constant.  Each
  combination was tested twice within a experiment, for a total of 10
  questions.

\end{itemize} 
The outputs were consistent with the explanation and the input 50\% of
the time, so subjects could not simply learn that the outputs were
always (in)consistent.  

Participants were recruited via Amazon Mechanical Turk.  Before
starting the experiment, they were given a tutorial on the
verification task and the interface.  Then they were given a set of
three practice questions.  Following the practice questions, they
started the core questions for each experiment.  They were told that
their primary goal was accuracy, and their secondary goal was speed.
While the questions were the same for all participants, the order of
the questions was randomized for every participant.  Each participant
only participated in one of the experiments.  For example, one
participant might have completed a 12-question experiment on the
effect of varying explanation length in the recipe domain, while
another would have completed 10-question experiment on the effect of
repeated terms in the clinical domain.  Experiments were kept short to
avoid subjects getting tired.

\paragraph{Metrics}
We recorded three metrics: response time, accuracy, and subjective
satisfaction.  Response time was measured as the number of seconds
from when the task was displayed until the subject hit the submit
button on the interface.  Accuracy was measured if the subject
correctly identified whether the output was consistent with the input
and the explanation (a radio button).  After each submitting their
answer for each question, the participant was also asked to
subjectively rate the quality of the explanation on a scale from one
to ten.

\begin{figure}[H]
  \centering
  \begin{subfigure}{.7\textwidth}
    \centering
    \includegraphics[width=\textwidth]{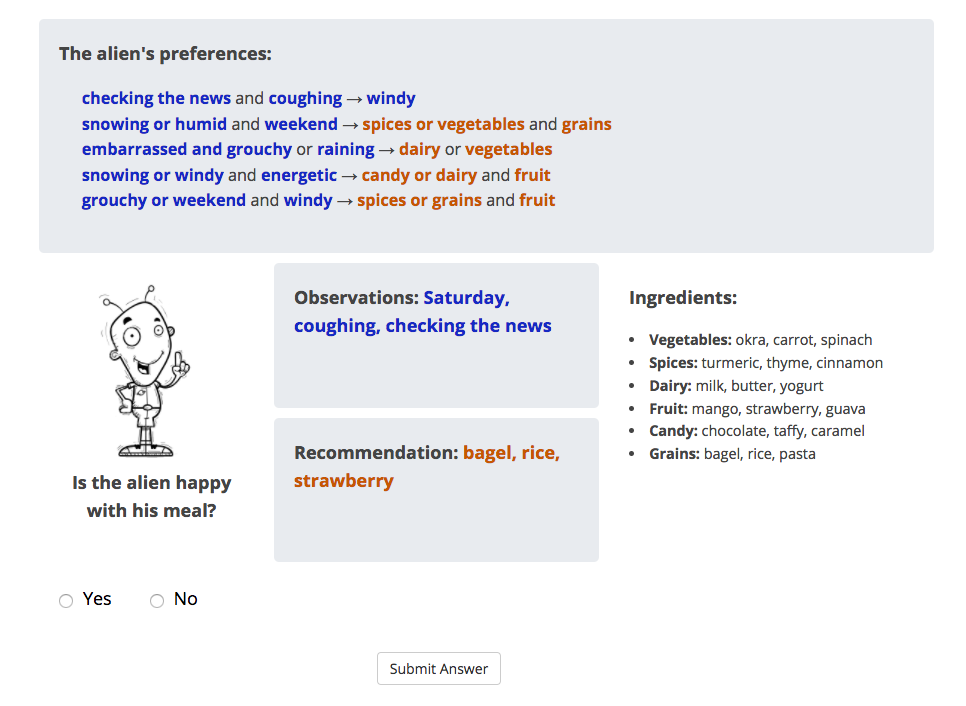}
    \caption{Recipe Domain}
  \end{subfigure}
  \begin{subfigure}{.7\textwidth}
    \centering  
    \includegraphics[width=\textwidth]{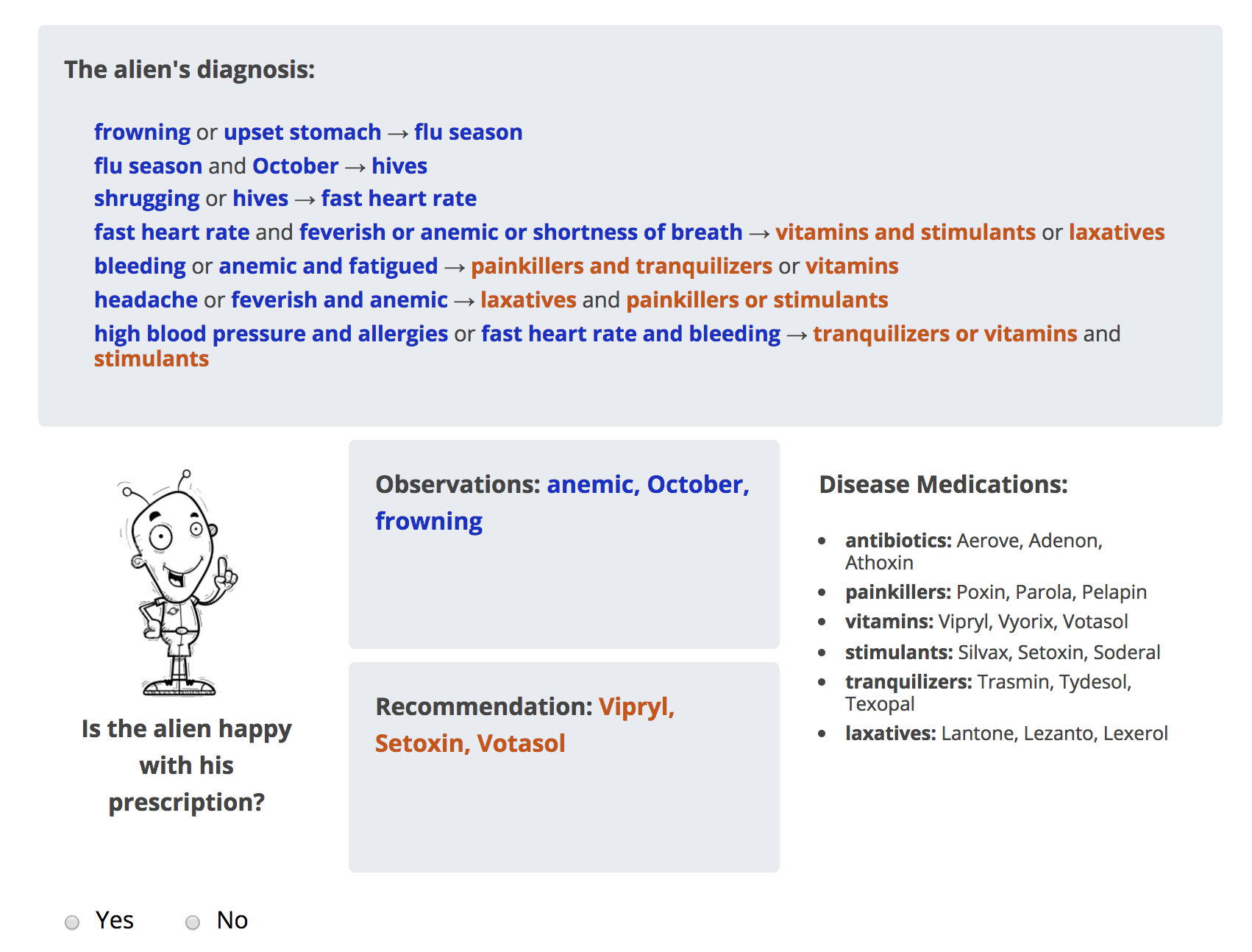}
    \caption{Clinical Domain}
  \end{subfigure}
  \caption{Screenshots of our interface for a task.  In the Recipe Domain, the supposed machine learning system has recommended the ingredients in the lower left box, based on its observations of the alien (center box).  The top box shows the system's explanation.  In this case, the recommended ingredients are consistent with the explanation and the inputs: The input conditions are weekend and windy (implied by coughing), and the recommendation of fruit and grains follows from the last line of the explanation.  In the Clinical Domain, the supposed machine learning system has recommended the medication in the lower box, based on its observations of the alien's symptoms (center box).  The top box shows the system's reasoning. The interface has exactly the same form as the recipe domain.}
  \label{fig:interface}
\end{figure}

\paragraph{Experimental Interface} 
Figure~\ref{fig:interface} shows our interfaces for the Recipe and
Clinical domains.  The \emph{observations} section (middle) refers to
the inputs into the algorithm.  The \emph{recommendation} section
refers to the output of the algorithm.  The \emph{preferences} section
(top) contains the explanation---the reasoning that the supposed
machine learning system used to suggest the output (i.e.,
recommendation) given the input, presented as a procedure in the form
of a decision set.  Finally, the \emph{ingredients} section in the
Recipe domain (and the \emph{disease medications} section in the
Clinical domain) contained a dictionary of \emph{cognitive chunks}
relevant to the experiment (for example, the fact that bagels, rice,
and pasta are all grains).  Including this list explicitly allowed us
to control for the fact that some human subjects may be more familiar
with various concepts than others.

The choice of location for these elements was chosen based on pilot
studies---while an ordering of input, explanation, output might make
more sense for an AI expert, we found that presenting the information
in the format of Figure~\ref{fig:interface} seemed to be easier for
subjects to follow in our preliminary explorations.  We also found
that presenting the decision set as a decision set seemed easier to
follow than converting it into paragraphs.  Finally, we colored the
input conditions in blue and outputs in orange within the
explanation. We found that this highlighting system made it easier for
participants to parse the explanations for input conditions.

\section{Results}
We recruited 100 subjects for each our six experiments, for a total of
600 subjects all together.  Table~\ref{tab:demographics} summarizes
the demographics of our subjects across the experiments.  Most
participants were from the US or Canada (with the remainder being almost
exclusively from Asia) and were less than 50 years old.  A majority
had a Bachelor's degree.  There were somewhat more male participants
than female.  We note that US and Canadian participants with moderate
to high education dominate this survey, and results may be different
for people from different cultures and backgrounds.  

\begin{table}
  \caption{Participant Demographics.  There were no patients over 69 years old.  4.2\% of participants reported ``other'' for their education level.  The rates of participants from Australia, Europe, Latin America, and South America were all less than 0.5\%.  (All participants were included in the analyses, but we do not list specific proportions for them for brevity.)} 
  \centering
  \begin{tabular}{|l||lll|}\hline
    Feature & Category : Proportion & & \\ \hline \hline 
    Age & 18-34 : 59.0\% & 35-50 : 35.1\% & 51-69 : 5.9\% \\ \hline
    Education & High School : 28.5\% & Bachelor's : 52.4\% & Beyond Bachelor's: 14.9\% \\ \hline
    Gender & Male : 58.8\% & Female : 41.2\% &  \\ \hline
    Region & US/Canada : 87.1\% & Asia : 11.4\% & \\ \hline
  \end{tabular}
  \label{tab:demographics}
\end{table}

All participants completed the full task (each survey was only 10-12
questions).  In the analysis below, however, we exclude participants
who did not get all three initial practice questions or all two of the
additional practice questions correct.  While this may have the effect
of artificially increasing the accuracy rates overall---we are only
including participants who could already perform the task to a
reasonable extent---this criterion helped filter the substantial
proportion of participants who were simply breezing through the
experiment to get their payment.  We also excluded one participant in
the clinical version of the cognitive chunks experiment who did get
sufficient practice questions correct but then took more than ten
minutes to answer a single question.  Table~\ref{tab:counts} describes
the total number of participants that remained in each experiment out
of the original 100 participants.

\begin{table}
  \caption{Number of participants who met our inclusion criteria for
    each experiment.}
  \centering
  \begin{tabular}{|l|l|l|}\hline
    & Recipe Domain & Clinical Domain \\ \hline
    Explanation Size & N=88 & N=73 \\ \hline
    New Cognitive Chunks & N=77 & N=73 \\ \hline
    Variable Repetition & N=70 & N=71 \\ \hline
  \end{tabular}
  \label{tab:counts} 
\end{table}

\begin{figure}[t]
\centering
\begin{tabular}{c|c|c}
\subcaptionbox{Recipe\_V1 Accuracy \label{2}}{\includegraphics[width = 2.10in]{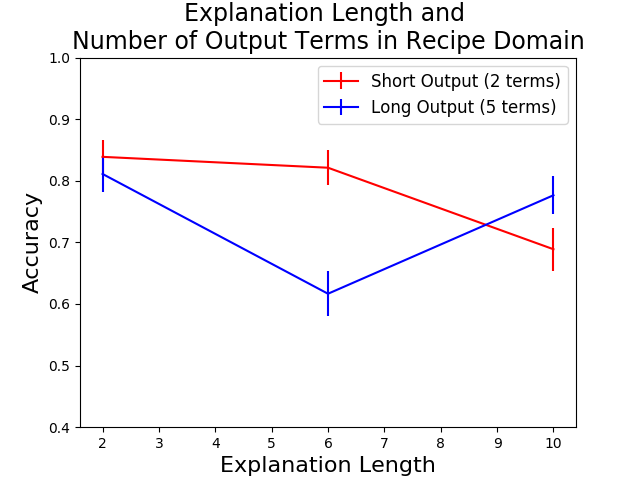}} &
\subcaptionbox{Recipe\_V2 Accuracy \label{1}}{\includegraphics[width = 2.10in]{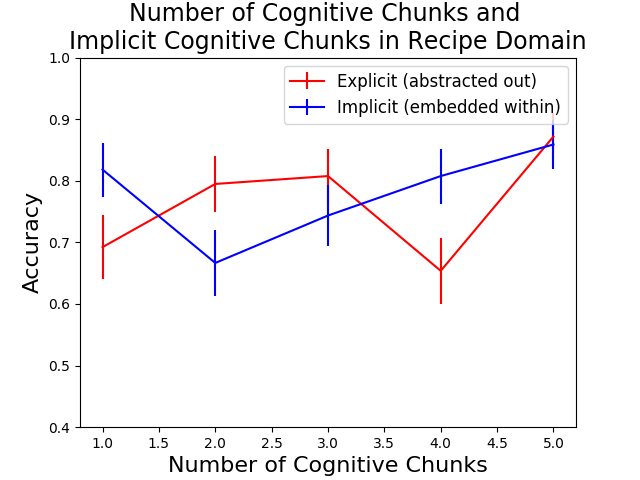}} &
\subcaptionbox{Recipe\_V3 Accuracy\label{1}}{\includegraphics[width = 2.10in]{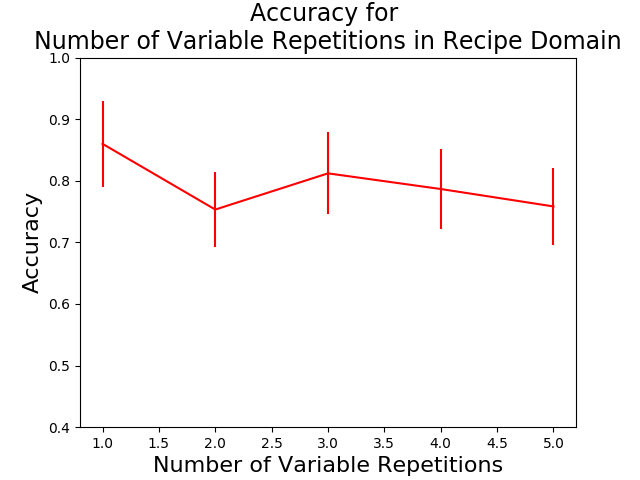}} \\
\midrule
\subcaptionbox{Clinical\_V1 Accuracy\label{2}}{\includegraphics[width = 2.10in]{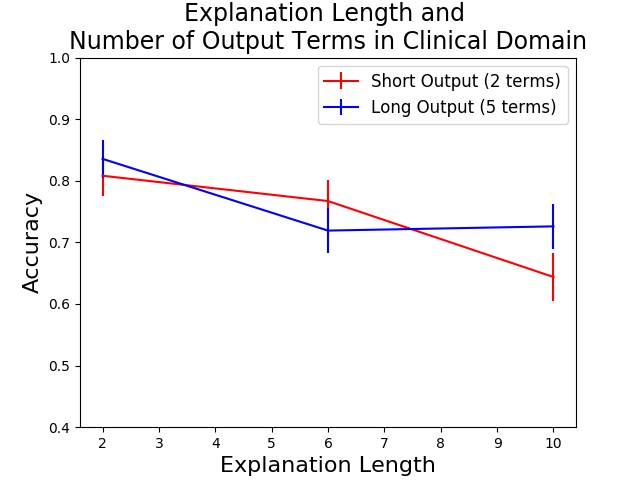}}  &
\subcaptionbox{Clinical\_V2 Accuracy\label{1}}{\includegraphics[width = 2.10in]{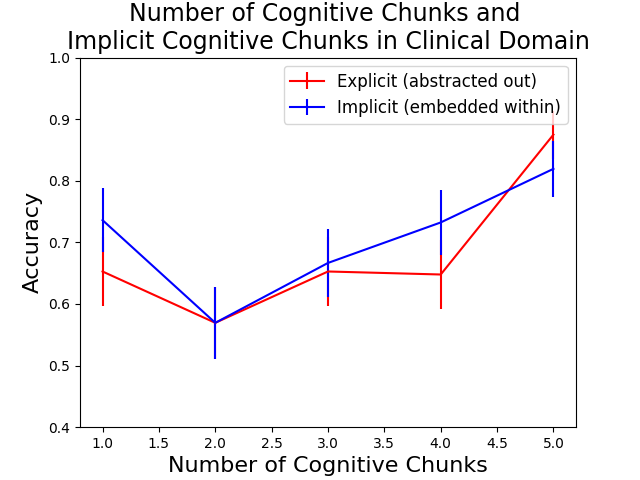}}  &
\subcaptionbox{Clinical\_V3 Accuracy\label{1}}{\includegraphics[width = 2.10in]{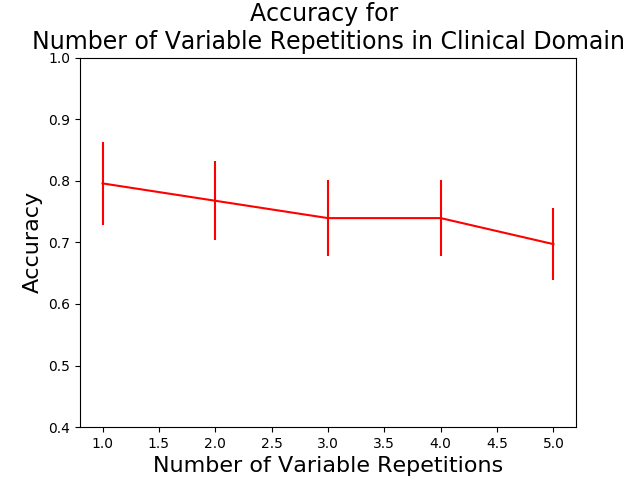}}  \\
\end{tabular}
\caption{Accuracy across the six experiments.  Vertical lines indicate standard errors.}
\label{fig:accuracy}
\end{figure}

\begin{figure}[t]
\centering
\begin{tabular}{c|c|c}
\subcaptionbox{Recipe\_V1 Time \label{2}}{\includegraphics[width = 2.10in]{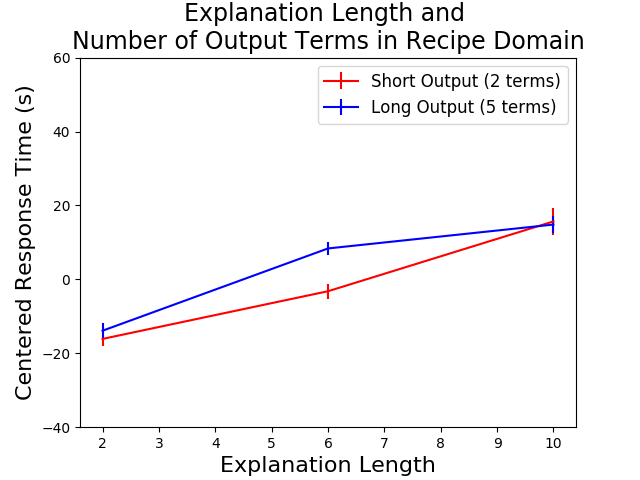}} &
\subcaptionbox{Recipe\_V2 Time \label{1}}{\includegraphics[width = 2.10in]{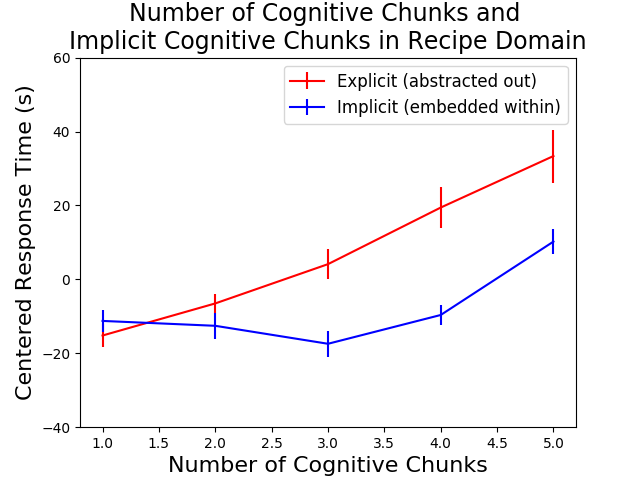}} &
\subcaptionbox{Recipe\_V3 Time\label{1}}{\includegraphics[width = 2.10in]{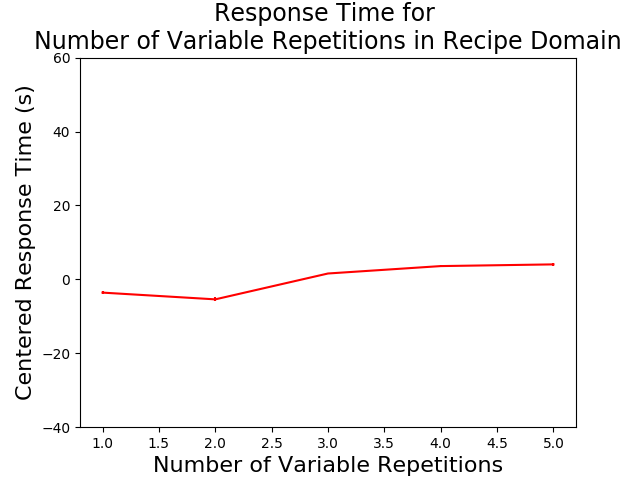}} \\
\midrule
\subcaptionbox{Clinical\_V1 Time\label{2}}{\includegraphics[width = 2.10in]{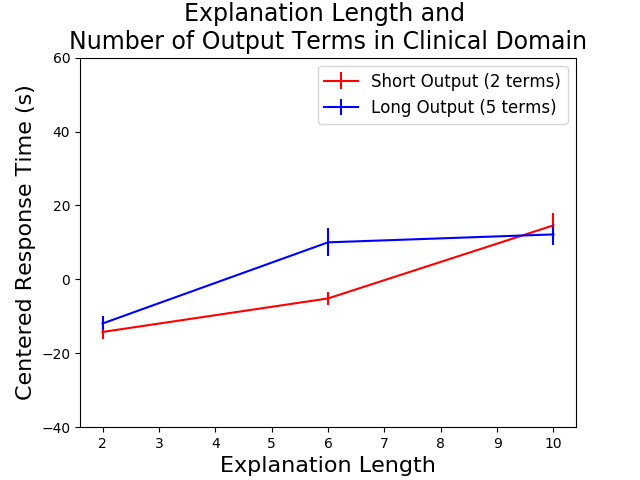}}  &
\subcaptionbox{Clinical\_V2 Time\label{1}}{\includegraphics[width = 2.10in]{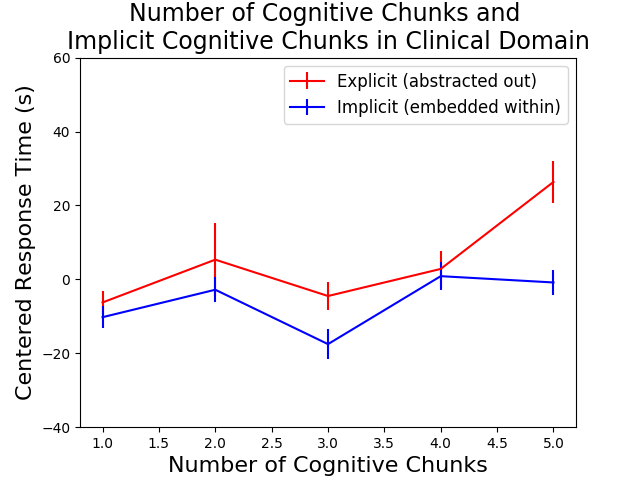}}  &
\subcaptionbox{Clinical\_V3 Time\label{1}}{\includegraphics[width = 2.10in]{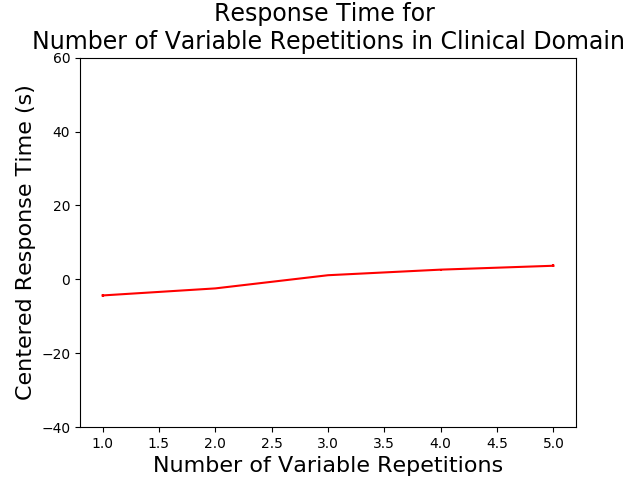}}  \\
\end{tabular}
\caption{Response times across the six experiments.  Responses were normalized by subtracting out subject mean to create centered values, and only response times for those subjects who got the question right are included.  Vertical lines indicate standard errors.}  
\label{fig:time}
\end{figure}

\begin{figure}[t]
\centering
\begin{tabular}{c|c|c}
\subcaptionbox{Recipe\_V1 Evaluation \label{2}}{\includegraphics[width = 2.10in]{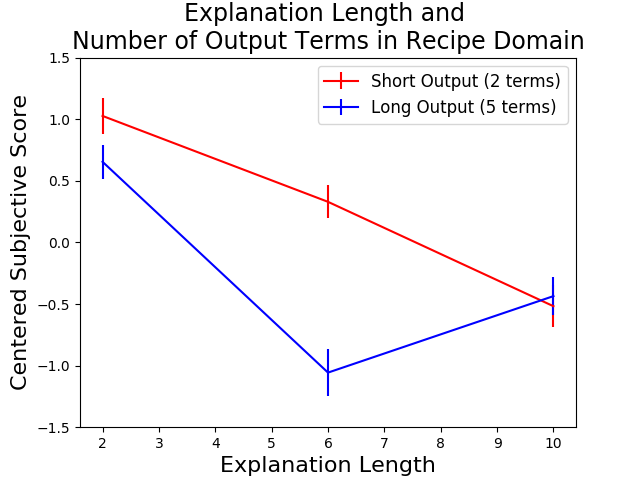}} &
\subcaptionbox{Recipe\_V2 Evaluation \label{1}}{\includegraphics[width = 2.10in]{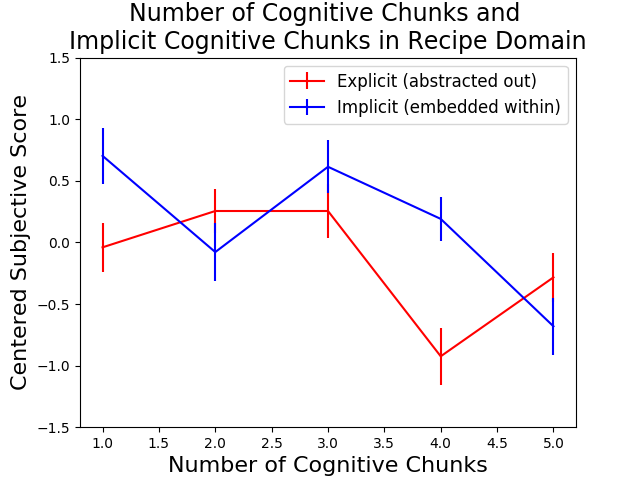}} &
\subcaptionbox{Recipe\_V3 Evaluation\label{1}}{\includegraphics[width = 2.10in]{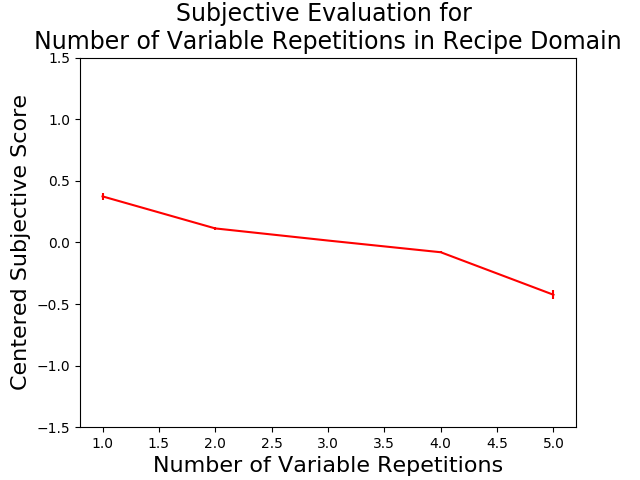}} \\
\midrule
\subcaptionbox{Clinical\_V1 Evaluation\label{2}}{\includegraphics[width = 2.10in]{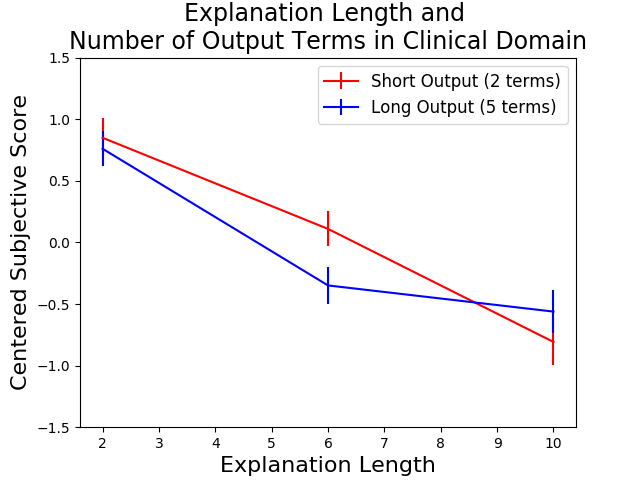}}  &
\subcaptionbox{Clinical\_V2 Evaluation\label{1}}{\includegraphics[width = 2.10in]{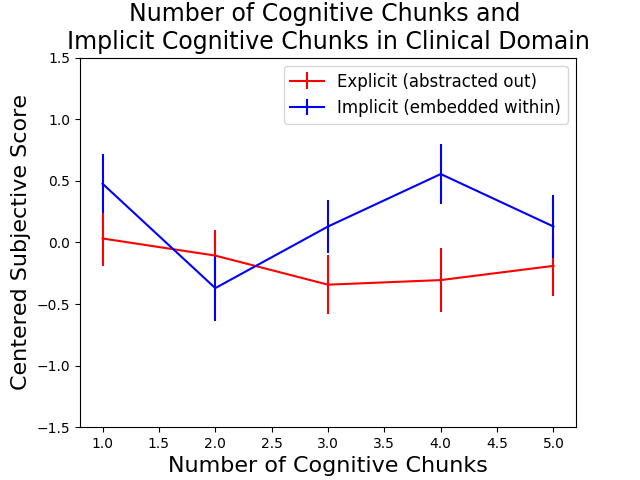}}  &
\subcaptionbox{Clinical\_V3 Evaluation\label{1}}{\includegraphics[width = 2.10in]{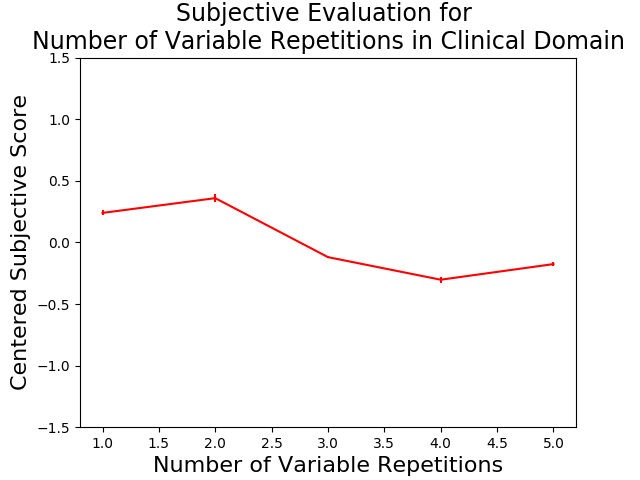}}  \\
\end{tabular}
\caption{Subjective evaluation of explanations in three experiments. Participants were asked to rate each explanation from 1 to 10. Responses were normalized by subtracting out subject mean to create centered values.}
\label{fig:subjective_evaluation}
\end{figure}

Figures~\ref{fig:accuracy} and~\ref{fig:time} present the accuracy and
response time across all six experiments, respectively.  Response time
is shown for subjects who correctly answered the questions.
Figure~\ref{fig:subjective_evaluation} shows the trends in the
participants' subjective evaluation---whether they thought the
explanation was easy to follow or not.  We evaluated the statistical
significance of the trends in these figures using a linear regression
for the continuous outputs (response time, subjective score) and a
logistic regression for binary outputs (accuracy).  For each outcome,
one regression was performed for each of the experiments V1, V2, and
V3.  If an experiment had more than one independent
variable---e.g. number of lines and terms in output---we performed one
regression with both variables.  Regressions were performed with the
statsmodels library \citep{seabold2010statsmodels} and included an
intercept term.  Table~\ref{tab:pvalues} summarizes these results.
  
\begin{table}
  \caption{Significance tests for each factor.  Linear and logistic
    regression weights were computed for continuous and binary outputs
    respectively (the subjective evaluations were treated as
    continuous).  A single regression was computed for each of V1, V2,
    and V3.  Highlighted p-values are significant at $\alpha$ = 0.05
    with a Bonferroni multiple comparisons correction across all tests
    of all experiments.}
  \centering
\begin{tabular}{|l|l|l|l|l|} \hline 
  \multicolumn{5}{c}{Accuracy} \\ \hline 
  & \multicolumn{2}{c}{Recipe} & \multicolumn{2}{|c|}{Clinical} \\ \hline
  Factor  & weight & p-value & weight & p-value \\ \hline

Explanation Length (V1) & -0.0116 & 0.00367 & -0.0171 & \textbf{0.000127} \\ \hline 
 Number of Output Terms (V1) & -0.0161 & 0.0629 & 0.00685 & 0.48\\ \hline 
 Number of Cognitive Chunks (V2) & 0.0221 & 0.0377 & 0.0427 & \textbf{0.00044} \\ \hline 
 Implicit Cognitive Chunks (V2) & 0.0147 & 0.625 & 0.0251 & 0.464\\ \hline 
 Number of Variable Repetitions (V3) & -0.017 & 0.104 & -0.0225 & 0.0506\\ \hline 


  \multicolumn{5}{c}{Response Time} \\ \hline 
  & \multicolumn{2}{c}{Recipe} & \multicolumn{2}{|c|}{Clinical} \\ \hline
  Factor  & weight & p-value & weight & p-value \\ \hline

Explanation Length (V1) & 3.77 & \textbf{2.24E-34} & 3.3 & \textbf{5.73E-22}\\ \hline 
 Number of Output Terms (V1) & 1.34 & 0.0399 & 1.68 & 0.0198\\ \hline 
 Number of Cognitive Chunks (V2) & 8.44 & \textbf{7.01E-18} & 4.6 & \textbf{1.71E-05}\\ \hline 
 Implicit Cognitive Chunks (V2) & -15.3 & \textbf{2.74E-08} & -11.8 & \textbf{0.000149} \\ \hline 
 Number of Variable Repetitions (V3) & 2.4 & \textbf{0.000659} & 2.13 & 0.0208\\ \hline 
  

  \multicolumn{5}{c}{Subjective Evaluation} \\ \hline 
  & \multicolumn{2}{c}{Recipe} & \multicolumn{2}{|c|}{Clinical} \\ \hline
  Factor  & weight & p-value & weight & p-value \\ \hline
Explanation Length (V1) & -0.165 & \textbf{5.86E-16} & -0.186 & \textbf{1.28E-19}\\ \hline 
 Number of Output Terms (V1) & -0.187 & \textbf{2.12E-05} & -0.0335 & 0.444\\ \hline 
 Number of Cognitive Chunks (V2) & -0.208 & \textbf{1.93E-05} & -0.0208 & 0.703\\ \hline 
 Implicit Cognitive Chunks (V2) & 0.297 & 0.0303 & 0.365 & 0.018\\ \hline 
 Number of Variable Repetitions (V3) & -0.179 & \textbf{5.71E-05} & -0.149 & \textbf{0.000771}\\ \hline 
  

\end{tabular}
\label{tab:pvalues}
\end{table}

\paragraph{In general, greater complexity results in higher response times and lower satisfaction.}   
Increasing the number of lines, the number of terms within a line,
adding new concepts, and repeating variables all increase the
complexity of an explanation in various ways.  In
figure~\ref{fig:time}, we see that all of these changes increase
response time.  The effect of adding lines to scan results in the
biggest increases in response time, while the effect of increasing the
number of variable repetitions is more subtle.  Making new concepts
explicit also consistently results in increased response time.  This
effect may have partly been because adding a new concept explicitly
adds a line, while adding it implicitly increases the number of terms
in a line---and from V1, we see that the effect of the number of lines
is larger than the effect of the number of terms.  However, overall
this effect seems larger than just adding lines (note the scales of
the plots).  Subjective scores appear to correlate inversely with
complexity and response time.

Perhaps most unexpected was that participants both took longer and
seemed less happy when new cognitive chunks were made explicit rather
than implicitly embedded in a line---we might have expected that even
if the explanation took longer to process, it would have been in some
senses easier to follow through. It would be interesting to unpack
this effect in future experiments, especially if participant
frustration came from there now being multiple relevant lines, rather
than just one. Future experiments could also highlight terms from the
input in the explanation, to make it easier for participants to find
the lines of potential relevance.  

\paragraph{Different explanation variations had little effect on accuracy.}
While there were strong, consistent effects from explanation variation
on response time and satisfaction, these variations seemed to have
much less effect on accuracy.  There existed general decreasing trends
in accuracy with respect to explanation length and the number of
variable repetitions, and potentially some increasing trends with the
number of new concepts introduced.  However, few were statistically
significant. This serves us as an interesting controlled comparison. In other words, we can now observe effects of different factors, holding the accuracy constant. The lack of effect may be because subjects were told to
perform the task as quickly as they could without making mistakes, and
our filtering also removed subjects prone to making many errors in the
practice questions.  Thus, the effect of a task being harder would be
to increase response time, rather than decrease accuracy.

The differences in direction of the trends---some increases in
complexity perhaps increasing accuracy, others decreasing it---are
consistent with findings that sometimes more complex tasks force
humans into slower but more careful thinking, while in other cases
increased complexity can lead to higher errors
\citep{kahneman2011thinking}.  Is it the case that the factors that
resulted in the largest increases in response time (new concepts) also
force the most concentration?  While these experiments cannot
differentiate these effects, future work to understand these phenomena
may help us identify what kinds of increased complexities in
explanation are innocuous or even useful and which push the limits of
our processing abilities.

\paragraph{Trends are consistent across recipe and clinical domains} 
In all experiments, the general trends in the metrics are consistent
across both the recipe and clinical domains. Sometimes an effect is
much weaker or unclear, but never is an effect clearly reversed.  We
believe this bodes well for there being a set of general principles
for guiding explanation design, just as there exist design principles
for interfaces and human-computer interaction. However, one small pattern can be noted in Figure 5, which shows lower satisfaction for the clinical domain than the recipe domain. This could be due to the fact that people felt more agitated about performing poorly in the medical domain than the clinical domain.

\section{Discussion and Conclusion}
Identifying how different factors affect a human's ability to utilize
explanation is an essential piece for creating interpretable machine
learning systems---we need to know what to optimize.  What factors
have the largest effect, and what kind of effect?  What factors have
relatively little effect?  Such knowledge can help us expand to
faithfulness of the explanation to what it is describing with minimal
sacrifices in human ability to process the explanation.

In this work, we investigated how the ability of humans to perform a
simple task---verifying whether an output is consistent with an
explanation and input---varies as a function of explanation size, new
types of cognitive chunks and repeated terms in the explanation.  We
tested across two domains, carefully controlling for everything but
the domain.

We summarized some intuitive and some counter intuitive discoveries---as any increase in explanation
complexity increases response time and decreases subjective
satisfaction with the explanation---some patterns were not so
obvious.  We had not expected that embedding a new concept would have
been faster to process and more appealing than creating a new
definition.  We also found that new concepts and the number of lines
increase response time more than variable repetition or longer lines.
It would be interesting to verify the magnitude of these sensitivities
on other tasks, such as forward simulation or counterfactual
reasoning, to start building a more complete picture of what we should
be optimizing our explanations for.  

More broadly, there are many interesting directions regarding what
kinds of explanation are best in what contexts.  Are there universals
that make for interpretable procedures, whether they be cast as
decision sets, decision trees, or more general pseudocode; whether the
task is verification, forward simulation, or counterfactual reasoning?
Do these universals also carry over to regression settings?  Or does
each scenario have its own set of requirements?  When the
dimensionality of an input gets very large, do trade-offs for defining
intermediate new concepts change?  A better understanding of these
questions is critical to design systems that can provide rationale to
human users.

\paragraph{Acknowledgements}
The authors acknowledge PAIR at Google and the Harvard Berkman Klein
Center for their support.

\bibliographystyle{plainnat}
\bibliography{main}

\appendix 

\section*{Description of Pilot Studies} 

We conducted several pilot studies in the design of these experiments.
Our pilot studies showed that asking subjects to respond quickly or
within a time limit resulted in much lower accuracies; subjects would
prefer to answer as time was running out rather than risk not
answering the question.  That said, there are clearly avenues of
adjusting the way in which subjects are coached to place them in fast
or careful thinking modes, to better identify which explanations are
best in each case.

The experiment interface design also played an important role.  We
experimented with different placements of various blocks, the coloring
of the text, whether the explanation was presented as rules or as
narrative paragraphs, and also, within rules, whether the input was
placed before or after the conclusion (that is, `if A: B'' vs. ``B if
A'').  All these affected response time and accuracy, and we picked
the configuration that had the highest accuracy and user satisfaction.

Finally, in these initial trials, we also varied more factors: number
of lines, input conjunctions, input disjunctions, output conjunctions,
output disjunctions and global variables. Upon running preliminary
regressions, we found that there was no significant difference in
effect between disjunctions and conjunctions, though the number of
lines, global variables, and general length of output
clause---regardless of whether that length came from conjunctions or
disjunctions---did have an effect on the response time. Thus, we chose
to run our experiments based on these factors.

\end{document}